\documentclass[pmlr,twocolumn,10pt]{jmlr}

\usepackage{rotating}
\usepackage{cprotect}
\usepackage{longtable}
\usepackage{booktabs}
\usepackage{siunitx}
\usepackage[switch]{lineno}
\usepackage{longtable}
\usepackage{mathtools}
\usepackage{dsfont}
\usepackage{soul}
\usepackage{algorithm}
\usepackage{pifont}
\usepackage{mdframed}
\usepackage{multirow}
\usepackage{comment}
\usepackage[shortlabels]{enumitem}

\definecolor{mygray}{gray}{0.8}
\DeclareMathOperator*{\argmax}{arg\,max}

\jmlrvolume{}
\jmlrpages{}
\jmlryear{}
\jmlrproceedings{}{}

\title[Pragmatic Policy Development via Interpretable Behavior Cloning]{Pragmatic Policy Development via Interpretable Behavior Cloning}

\author{%
\Name{Anton Matsson} \Email{antmats@chalmers.se} \\
\addr Chalmers University of Technology and University of Gothenburg
\AND
\Name{Yaochen Rao} \Email{yaochenr@chalmers.se} \\
\addr Chalmers University of Technology and University of Gothenburg
\AND
\Name{Heather J. Litman} \Email{heather.litman@thermofisher.com} \\
\addr Thermo Fisher Scientific
\AND
\Name{Fredrik D. Johansson} \Email{fredrik.johansson@chalmers.se} \\
\addr Chalmers University of Technology and University of Gothenburg
}

\begin{document}

\maketitle

\begin{abstract}
Offline reinforcement learning (RL) holds great promise for deriving optimal policies from observational data, but challenges related to interpretability and evaluation limit its practical use in safety-critical domains. Interpretability is hindered by the black-box nature of unconstrained RL policies, while evaluation—typically performed off-policy—is sensitive to large deviations from the data-collecting behavior policy, especially when using methods based on importance sampling. To address these challenges, we propose a simple yet practical alternative: deriving treatment policies from the most frequently chosen actions in each patient state, as estimated by an interpretable model of the behavior policy. By using a tree-based model, which is specifically designed to exploit patterns in the data, we obtain a natural grouping of states with respect to treatment. The tree structure ensures interpretability by design, while varying the number of actions considered controls the degree of overlap with the behavior policy, enabling reliable off-policy evaluation. This pragmatic approach to policy development standardizes frequent treatment patterns, capturing the collective clinical judgment embedded in the data. Using real-world examples in rheumatoid arthritis and sepsis care, we demonstrate that policies derived under this framework can outperform current practice, offering interpretable alternatives to those obtained via offline RL.
\end{abstract}


\section{Introduction}
\label{sec:introduction}

Observational data provide a valuable foundation for applying machine learning (ML) to derive treatment policies that enhance clinical decision-making~\citep{chakraborty2013statistical}. Recently, significant attention has been devoted to reinforcement learning (RL), a branch of ML focused on learning optimal decision-making strategies~\citep{sutton2018rl}. While classical RL relies on trial-and-error learning—ill-suited to high-stakes domains like healthcare—offline RL learns from previously collected data, offering the potential to turn static datasets into effective decision-making engines~\citep{levine2020offline}. However, applying offline RL in clinical settings presents several well-known challenges~\citep{yu2021reinforcement,jayaraman2024primer}, with the lack of interpretability and the difficulty of policy evaluation being among the most significant. These limitations raise the question of whether other approaches may be better suited for practical clinical use.

Policy evaluation involves assessing the performance of a newly derived policy, often referred to as the target policy. In safety-critical domains, this evaluation—like policy learning—must be performed using an offline dataset, a problem known as off-policy evaluation (OPE). A large class of OPE methods rely on importance sampling (IS)~\citep{precup2000eligibility}, where outcomes from patient trajectories collected under the behavior policy—that is, the current, observed decision-making behavior—are weighted by how likely those trajectories would be under the target policy. When the target and behavior policies differ substantially, IS-based estimates of performance tend to exhibit high variance. While this issue can be mitigated—for example, by normalizing weights~\citep{precup2000eligibility} or incorporating model-based components into the estimator~\citep{jiang2016doubly,thomas2016data,farajtabar2018more}—reliable OPE generally requires that the policies are sufficiently similar~\citep{gottesman2018evaluating,voloshin2021empirical}. 

The challenge of interpretability arises from the fact that much of RL's recent success is due to its integration with deep learning (deep RL), where black-box neural networks are used to represent policies~\citep{mnih2015human}. Such target policies are typically difficult to interpret, and this lack of transparency can prevent domain experts from identifying errors or artifacts, potentially undermining trust in medical applications~\citep{pace2022poetree,lipton2017doctor}. While there have been efforts to improve the interpretability of RL policies—either directly by defining interpretable policy classes~\citep{silva2020optimization,verma2019imitation,hein2018interpretable}, or indirectly by distilling interpretable policies from black-box models~\citep{verma2018programmatically,liu2018toward}—the prevailing view is that deep RL is not yet ready for deployment in high-stakes domains such as healthcare~\citep{glanois2024survey}.

Combining ideas from interpretable RL with robust offline RL, where the target policy is constrained to stay close to the behavior policy to improve evaluability~\citep{fujimoto2019off,kostrikovoffline,kumar2020conservative}, offers a promising direction for addressing these practical limitations. However, before deploying the full RL machinery, it is reasonable to ask: can we develop interpretable policies that are amenable to reliable OPE in a simpler and more pragmatic way? 

\paragraph{Contributions}
We propose using supervised learning of the behavior policy—also known as behavior cloning~\citep{torabi2018behavioral}—to derive interpretable and evaluable target policies for clinical decision-making. Specifically, the proposed policy is constructed based on the most frequently chosen treatments in each patient state, as estimated by the behavior policy model, with the option to incorporate their observed outcomes to further guide treatment selection. As such, the policy exploits the collective clinical expertise embedded in the data. By varying the number of treatments considered, we control the degree of overlap with the behavior policy, facilitating reliable OPE. Furthermore, by choosing an interpretable model class for the behavior policy, the resulting target policy is interpretable by design. While several model classes are possible, we recommend using a tree-based structure, as it provides a natural partitioning of the patient space based on observed treatment patterns. To utilize a common situation in clinical decision-making—where patients often remain on the same treatment across decision points—we construct a meta-model that uses separate trees to predict whether a patient will switch treatments and, if so, which treatment they will switch to.

We refer to this approach as \textit{pragmatic policy development}. While we cannot guarantee that such policies outperform current practice, they are explicitly designed to be amenable to OPE, enabling meaningful comparison. In experiments, we find that policies developed under this framework are, on average, estimated to have higher policy values than current practice in real-world examples from the management of rheumatoid arthritis (RA) and sepsis. In contrast, policies derived using offline RL yield estimates with high variance, raising questions about their practical relevance.

\section{Leveraging Observational Data to Improve Clinical Decision-Making}
\label{sec:seq_decisions}

The treatment of patients with chronic or acute diseases, such as RA and sepsis, can be formulated as a multi-stage decision process involving states $S_t\in\mathcal{S}$, actions $A_t\in\{1, \ldots, K\}$, and rewards $R_t\in{\mathbb{R}}$. At each stage $t=1, \ldots, T$, the clinician selects a treatment $A_t$ based on the patient's medical history, summarized in the state $S_t$; the reward $R_t$ reflects the outcome of the chosen treatment. A patient's medical history comprises the sequence of covariates, actions, and rewards observed up to stage $t$, where covariates $X_t\in\mathcal{X}$ include, for example, demographics, diagnostic test results, and comorbidities. Prior work by \cite{matsson2024how} suggests that a sufficient state representation includes the current covariates $X_t$, the most recent action $A_{t-1}$, and the most recent reward $R_{t-1}$, along with simple aggregates of earlier history.

The process by which clinicians treat patients gives rise to trajectories of state-action-reward triplets, $\tau=S_1, A_1, R_1, \ldots, S_{T}, A_{T}, R_{T}$, which are assumed to be collected in a dataset $\mathcal{D}$. Patterns in how clinicians select treatments define the \emph{behavior policy} $\mu$. We quantify these treatment patterns by estimating $p_{\mu}(A_t \mid S_t)$, the probability of selecting treatment $A_t$ given state $S_t$ under current clinical practice, based on state-action pairs in $\mathcal{D}$. In other contexts, this is often referred to as propensity estimation~\citep{abadie2016matching}, policy recovery~\citep{deuschel2024contextualized} or behavior cloning~\citep{torabi2018behavioral}. We let $p_{\hat{\mu}}(A_t \mid S_t)$, or simply $\hat{\mu}$, denote a probabilistic model of the behavior policy. 
 
A \emph{target policy} $\pi$ represents an alternative treatment strategy. We use $p_\pi(A_t \mid S_t)$ to denote the probability of taking action $A_t$ in state $S_t$ under this policy. Reinforcement learning is a classical approach to optimizing such policies~\citep{sutton2018rl}, where optimality is defined in terms of the expected cumulative reward, or value, $V^\pi$: $V^\pi \coloneqq \mathbb{E}_\pi\left[\sum_{t=1}^{T} R_{t} \right]$. In a clinical context, the target policy must typically be derived solely from observations in $\mathcal{D}$---a setting referred to as offline RL (see, for example, \citet{levine2020offline} for an overview). In principle, any off-policy RL algorithm, where the policy being learned differs from the data-collecting policy, can be applied to the fixed dataset $\mathcal{D}$. Most prior works in clinical settings have used methods based on Q-learning~\citep{yu2021reinforcement}.

\subsection{The Need for Interpretability and Evaluability}

Before a target policy $\pi$ can be deployed in clinical practice, its performance must be assessed relative to the behavior policy $\mu$. In the offline setting, where no data collected under the target policy is available, this evaluation amounts to estimating the value of $\pi$ using data collected under $\mu$—a problem known as off-policy evaluation. OPE is inherently challenging due to the difference between the two policies, and the difficulty increases as the divergence between $\pi$ and $\mu$ grows. This issue is especially pronounced in methods based on importance sampling~\citep{precup2000eligibility}, where observed rewards are weighted by the importance weights $\prod_{t}\frac{p_\pi(a_t \mid s_t)}{p_{\hat{\mu}}(a_t \mid s_t)}$, but exists for other methods as well. For example, value-based~\citep{le2019batch} or model-based~\citep{mythesis} approaches require extrapolation to account for the mismatch between the policies. As a result, {the target policy must remain sufficiently close to the behavior policy} to enable reliable offline evaluation~\citep{gottesman2018evaluating}.

A large class of offline RL methods tackles the distributional shift between target and behavior policies—which poses challenges not only for evaluation but also during training and deployment~\citep{levine2020offline}—by explicitly constraining the target policy to remain close to the behavior policy during learning~\citep{fujimoto2019off,kostrikovoffline,kumar2020conservative}. However, a key drawback of these methods—and RL algorithms more broadly—is their reliance on black-box neural networks, which limits their transparency to end users. For example, \citet{raghu2017deep} learned a sepsis treatment policy using deep RL that recommended lower doses of vasopressors for severely ill patients compared to clinicians. The black-box nature of this policy made it difficult to interpret the rationale behind such recommendations. This level of opacity would be unacceptable in clinical practice, where interpretability is often considered a prerequisite for the effective implementation of machine learning~\citep{stiglic2020interpretability}.

Interpretable RL is an emerging field, with existing approaches exploring, for example, decision trees to represent the target policy~\citep{roth2019conservative,silva2020optimization}. However, most work in this area relies on post-hoc explanations of policies represented by neural networks~\citep{verma2018programmatically,coppens2019distilling,bastani2018verifiable}, in contrast to the preferable approach of constructing directly interpretable policies~\citep{rudin2019stop}. Moreover, few of these methods are specifically designed for the offline setting, underscoring the need for best practices in developing interpretable and reliably evaluable policies for clinical decision-making.

\section{Pragmatic Policy Development via Interpretable Behavior Cloning}
\label{sec:framework}

Motivated by the need for interpretability and evaluability, we propose a pragmatic approach to policy development based on interpretable modeling of the behavior policy. An interpretable behavior policy model enables explanation of current decision-making behavior~\citep{pace2022poetree} and provides insight into which types of policies can be reliably evaluated with statistical support~\citep{matsson2022case}. Additionally, it facilitates reasoning about whether the state $S_t$ accounts for confounding variables that causally affect both the treatment decision $A_t$ and its outcome $R_t$. The absence of unmeasured confounding, along with overlapping support (that is, $p_{\mu}(a \mid s) > 0$ whenever $p_{\pi}(a \mid s) > 0$), are necessary assumptions for IS-based OPE~\citep{namkoong2020off}.

A target policy is derived from the most frequently chosen treatments in each state, as estimated by the behavior policy model. The most direct form of this approach, which is generalized and extended in Section~\ref{sec:extension}, defines the policy to always choose the most common treatment; that is, setting $p_{\pi}(A_t = a_t \mid S_t = s_t)$ to 1 if $a_t = \arg\max_{a} p_{\hat{\mu}}(a \mid S_t = s_t)$, and 0 otherwise. This yields a deterministic target policy that closely resembles the behavior policy as estimated by the model. Implementing such a policy in practice can be viewed as standardizing common treatment patterns, leveraging the collective expertise of clinicians represented in the data. Since the behavior policy model is interpretable, the resulting target policy is interpretable by design. From an OPE perspective, this construction ensures overlapping support with the behavior policy—that is, any action recommended by the target policy has nonzero probability under the behavior policy.

For the interpretable behavior policy model, we utilize decision trees for probabilistic classification. A decision tree partitions inputs into homogeneous groups, each representing a terminal (leaf) node. By fitting a decision tree to observed state-action pairs $(s_t, a_t)$, each leaf node groups patients with similar propensity scores $p_\mu(A_t \mid S_t)$. Matching subjects on the propensity score is sufficient to adjust for confounding~\citep{rosenbaum1983central}.
If the state $S_t$ fully captures all confounding variables, the variation in $A_t$ within each leaf of a sufficiently deep tree reflects practice variation among clinicians that does not stem from confounding.

Tree-based models are often used to identify subgroups within a population~\citep{keramati2022identification}, and behavior cloning with decision trees has been applied in domains such as robotic control~\citep{cichosz2014imitation,sheh2011behavioural}. However, other classes of interpretable models can also be employed. For example, prototypical networks~\citep{li2018deep,ming2019interpretable}, a form of case-based reasoning, perform a soft clustering of the state–action space when used for behavior policy modeling~\citep{matsson2022case}. The key desideratum is a model that naturally groups the data. As such, (generalized) linear models—while interpretable—are less suitable, as they do not inherently induce such groupings.

\subsection{Generalizing and Extending the Framework}
\label{sec:extension}

We recommend that researchers always consider the deterministic target policy for evaluation. However, depending on the amount and sparsity of the data, it may not be possible to evaluate such a policy with statistical confidence. In the general case, the target policy can be constructed based on the set of the $k$ actions with the highest probability in state $s_t$ under the behavior policy model $\hat{\mu}$, denoted as $\mbox{Top-}k(s_t; \hat{\mu})$. Formally, let
\begin{equation}
p_{\pi}(a_t \mid s_t) \coloneqq
\left\{
    \begin{array}{ll}
        \frac{p_{\hat{\mu}}\left(a_t \mid s_t\right)}{Z_k}, & \mbox{if } a_t \in \mbox{Top-}k(s_t; \hat{\mu});  \\
        0, & \mbox{otherwise},
    \end{array}
    \right.
    \label{eq:target}
\end{equation}
where the normalization constant $Z_k = \sum_{a \in \mbox{Top-}k(s_t; \hat{\mu})}p_{\hat{\mu}}(a \mid s_t)$ ensures that $p_{\pi}$ defines a valid probability distribution. By setting $k>1$, we obtain a stochastic policy that more closely resembles the behavior policy as $k$ approaches $K$, thereby increasing the effective sample size in evaluation~\citep{mcbook}. In practice, implementing a target policy with $k>1$ requires the clinician to choose which of the $k$ treatment alternatives to administer to the patient. In this case, the probabilities defined in Equation~\eqref{eq:target} can help guide this decision.

We can naturally extend Equation~\eqref{eq:target} to incorporate observed outcomes. Specifically, for each leaf in the fitted tree, we compute the average outcome for each action. The target policy is then defined by selecting the treatment---among the $k$ most common ones---that is associated with the highest outcome. Formally, let $O(s, a; \hat{\mu})$ map the state-action pair $(s, a)$ to the average outcome observed for patients in the same leaf as  $s$ who received $a$, as determined by the (tree-based) behavior policy model $\hat{\mu}$. In state $s_t$, the outcome-guided target policy selects the action
\begin{equation}
\argmax_{a\in\mbox{Top-}k(s_t; \hat{\mu})}O(s_t,a;\hat\mu).
\label{eq:target_outcomes}
\end{equation}

\paragraph{A Note on Terminology}
We refer to the general approach as pragmatic policy development via interpretable behavior cloning to reflect its reliance on supervised learning of observed treatment patterns. However, the policies defined in Equations~\eqref{eq:target} and~\eqref{eq:target_outcomes} differ from classical behavior cloning, which seeks to exactly replicate observed actions in each state—similar to the deterministic target policy introduced first in this section. A key assumption that distinguishes our approach from standard behavior cloning is that the dataset $\mathcal{D}$ contains examples from multiple practitioners, enabling us to exploit the collective knowledge embedded in the data.

\section{Exploiting Known Structure to Improve Modeling}
\label{sec:structure}

\begin{figure}[t]
\floatconts
  {fig:dt_only}
  {\caption{The topmost nodes of a standard decision tree fitted to a cohort of RA patients. The left side of the tree is dominated by the previous-treatment variable, $A_{t-1}$, leading to distinct subtrees for different treatment types. This structure results in repeated decision rules across subtrees, contributing to unnecessary model complexity.}}
  {\includegraphics[width=\linewidth]{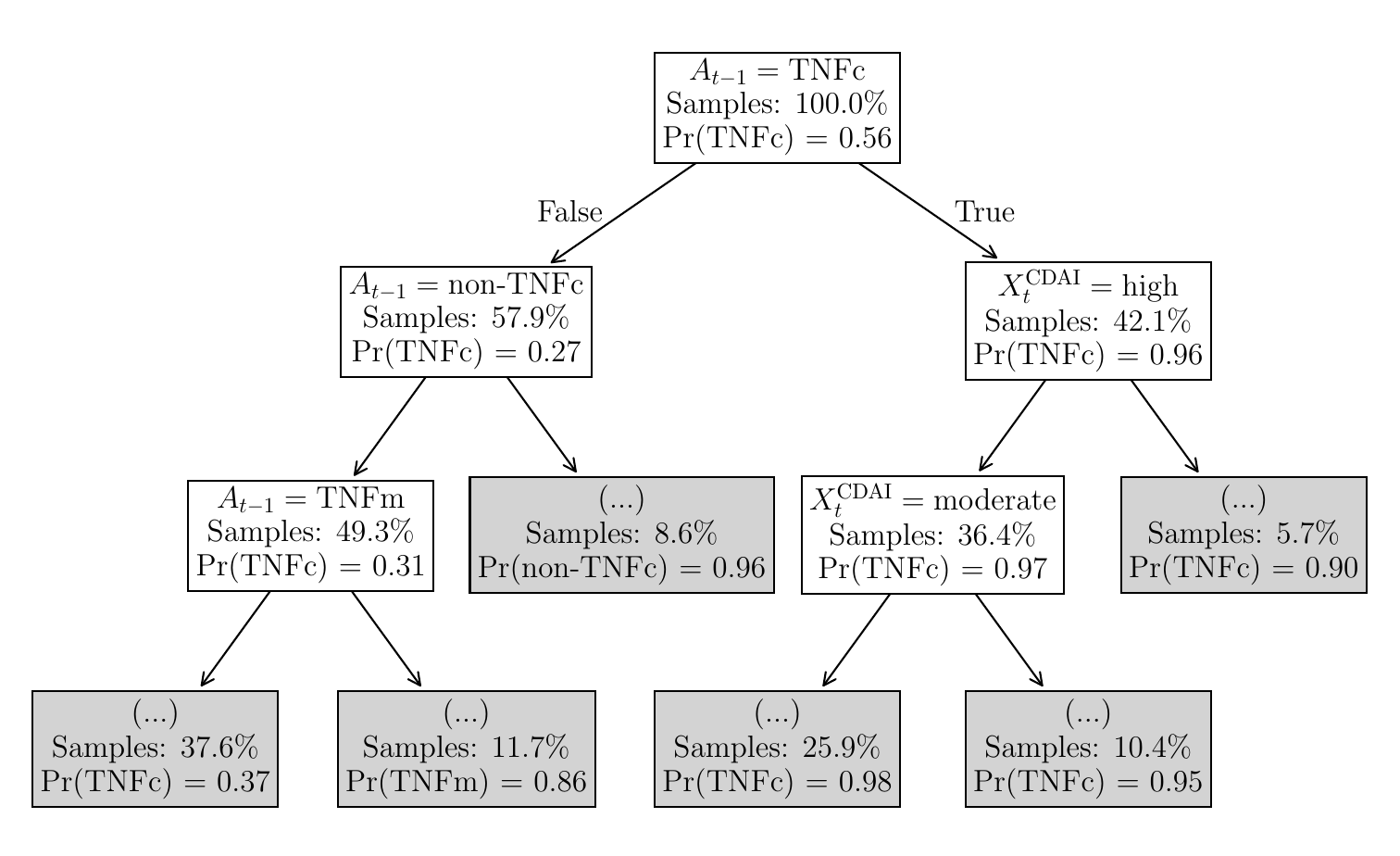}}
\end{figure}

In clinical decision-making—particularly for chronic diseases—it is common for patients to receive the same treatment across multiple decision points. For example, if a treatment is effective, there is typically no need to change it. This presents a challenge when modeling the behavior policy using a decision tree: the model is often dominated by the tendency to continue with the previous treatment. As illustrated in Figure~\ref{fig:dt_only} for the case of RA management, the previous-treatment variable, $A_{t-1}$, dominates the left branch of the tree, effectively partitioning it into distinct subtrees for each treatment type. Many of the same decision rules are repeated across subtrees, leading to unnecessary model complexity.

We leverage this structure in the data to improve both the accuracy and interpretability of the behavior policy model. Specifically, we construct a meta-estimator that combines two decision tree classifiers: (1) a binary classifier that predicts whether a patient switches treatments, and (2) a multi-class classifier that predicts which treatment is chosen, trained only on treatment switch events. Let $p_{\hat{\mu}}^{s}(S_t)$ and $p_{\hat{\mu}}^{t}(S_t)$ denote the probabilistic output of classifier (1) and (2), respectively. Furthermore, let $C_t$ be a binary random variable, where $C_t=1$ represents a treatment change, and $C_t=0$ indicates continuation of the current treatment. 

Formally, we define $p_{\hat{\mu}}^{s}(S_t) \coloneqq p_{\hat{\mu}}(C_t=1 \mid S_t)$ and $p_{\hat{\mu}}^{t}(k \mid S_t) \coloneqq p_{\hat{\mu}}(A_t=k \mid S_t)$. Let $\tilde{p}_{\hat{\mu}}^{t}(k \mid S_t)$ denote the probability of selecting treatment $k$ in state $S_t$ given that a treatment change occurs, $\tilde{p}_{\hat{\mu}}^{t}(k \mid S_t) \coloneqq p_{\hat{\mu}}(A_t=k \mid C_t=1, S_t)$. We obtain this probability by excluding the probability of the previous treatment:
\begin{equation*}
    \tilde{p}_{\hat{\mu}}^{t}(k \mid S_t) = \frac{
    \mathds{1}[k \neq a_{t-1}]
    p_{\hat{\mu}}^{t}(k \mid S_t)
    }{
    \sum_{j}
    \mathds{1}[j \neq a_{t-1}]
    p_{\hat{\mu}}^{t}(j \mid S_t)
    }.
\end{equation*}
Finally, we obtain the probability $p_{\hat{\mu}}(A_t=k \mid S_t)$, which is the output of the meta-estimator, by incorporating the probability of staying on the same treatment:
\begin{align}
p_{\hat{\mu}}(A_t = k \mid S_t) 
&= \left(1 - p_{\hat{\mu}}^{s}(S_t)\right) \cdot \mathds{1}[k = a_{t-1}] \nonumber \\
&\quad + p_{\hat{\mu}}^{s}(S_t) \cdot \tilde{p}_{\hat{\mu}}^{t}(S_t, k).
\label{eq:meta_estimator}
\end{align}

We use the meta-estimator \eqref{eq:meta_estimator} with decision trees for the switch and treatment classifiers to model the behavior policy in our experiments. We construct target policy candidates based on the most common treatments predicted by the model, as discussed in the previous section. To incorporate outcomes into the target policy, we compute the average observed outcome for each action within each leaf of the two trees. Let the state $s$ fall into leaf $i$ of the switch tree and leaf $j$ of the treatment tree. For this state, the outcome associated with the previous treatment is the average outcome among patients in leaf $i$ of the switch tree who remained on the same treatment ($c = 0$). The outcomes for other treatments are given by the average outcome observed among patients who switched treatments ($c = 1$) and belong to leaf $j$ of the treatment tree.

Using a separate model to estimate the probability that a patient switches treatment in a given state $s$ allows for adjusting the switching probability when constructing target policy candidates within the pragmatic policy development framework. We elaborate on this idea in Appendix~\ref{app:switch_adjustment}.

\section{Experiments}

We demonstrate our proposed approach---deriving candidate target policies based on an interpretable behavior policy model---using two real-world clinical examples: the management of RA and sepsis. We begin by modeling the behavior policy to evaluate the structured learning approach introduced in the previous section. We then construct target policies from these models and assess their performance relative to those learned via RL.

\paragraph{RA}
Using data from the PPD\texttrademark{} CorEvitas\texttrademark{} RA registry~\citep{corevitas}, we study RA treatment beginning with the initiation of the first biologic or targeted synthetic disease-modifying antirheumatic drug (b/tsDMARD). We identify a cohort of 1,565 patients who (i) have no history of b/tsDMARD use before registry enrollment, (ii) initiate at least one b/tsDMARD during registry participation, and (iii) have at least 2 years (720 days) of registry follow-up. As detailed in Appendix~\ref{app:data}, we allow the use of excluded data for behavior policy modeling. To simplify the action space, we group DMARDs into classes, resulting in eight distinct treatment options. We define the reward function as $R_t\coloneqq10-I_{t+1}$, where $I$ is the clinical disease activity index (CDAI). A CDAI of 10 marks the threshold between low and moderate-to-high disease activity.

\paragraph{Sepsis}
We use code provided by \citet{komorowski2018artificial} to collect and preprocess data for 11,482 patients meeting the international Sepsis-3 criteria~\citep{singer2016third} from the MIMIC-III database~\citep{johnson2016mimic}. For each patient, we include up to 24 hours of measurements starting from the estimated onset of sepsis. Data are discretized into 4-hour time steps, with multiple measurements aggregated within each interval. Following \citet{luo2024position}, we exclude patients with inconsistent time-series data. The action space includes 25 discrete actions, formed by combining doses of intravenous fluids and vasopressors, each binned into 5 levels. The reward is zero at all time steps except the final one, which is +100 if the patient survives and $-100$ otherwise.

\paragraph{Experimental Setup}
We split the data into separate training and testing sets. The training set is used to model the behavior policy and construct candidate target policies, while the testing set is used for OPE. A portion (\SI{20}{\percent}) of the training data is reserved for model validation and calibration. The entire procedure is repeated 50 times with different random seeds. Further details, including model architectures and hyperparameter selection, are provided in the following sections and Appendix~\ref{app:exp_details}.

\subsection{Behavior Policy Modeling}
\label{sec:mu_modeling}

For behavior policy modeling, we include three types of tree-based models: (i) a standard decision tree (DT); (ii) the meta-model described in Section~\ref{sec:structure} (DT-S), which uses two decision trees to separately predict treatment switching and treatment assignment; and (iii) a variant (DT-BLS) that combines DT-S with a separate decision tree for treatment classification at the first time step (baseline). DT-BLS accounts for the fact that baseline treatment decisions are often more predictable than those made later. For example, in RA, most patients initiate their first b/tsDMARD therapy with a Tumor Necrosis Factor (TNF) inhibitor~\citep{smolen2023eular}. Finally, we include a recurrent neural network (RNN) model as a baseline to assess how well the behavior policy can be estimated when using the full patient history. For the tree-based models, a patient’s state is represented using a combination of the most recent covariates and selected aspects of their history.

\begin{table}[t!]
\small
\centering
\caption{Average test AUROC and SCE for different models of the behavior policy across 50 splits of each dataset. We compare a standard decision tree (DT) with the model described in Section~\ref{sec:structure}, where a separate decision tree is used to predict treatment switching (DT-S). Additionally, we include a model that combines DT-S with a decision tree for classification at the first time step or baseline (DT-BLS).
}
\label{tab:scores}
\begin{tabular}{lllll}
\toprule
 & \multicolumn{2}{c}{\textbf{RA}} & \multicolumn{2}{c}{\textbf{Sepsis}} \\
 \cmidrule(lr){2-3} \cmidrule(lr){4-5}
Model & AUROC & SCE  & AUROC  & SCE  \\
\midrule
DT & 92.0 & 2.7 & 86.9 & 0.4 \\
DT-S & 92.8 & 2.6 & 86.0 & 0.5 \\
DT-BLS & 94.9 & 1.3 & 86.8 & 0.5 \\
RNN & 91.8 & 2.4 & 88.1 & 0.5 \\
\bottomrule
\end{tabular}
\end{table}

A good behavior policy model should achieve strong predictive performance while also producing well-calibrated probabilities. In Table~\ref{tab:scores}, we report the average area under the receiver operating characteristic curve (AUROC) and static calibration error (SCE)~\citep{nixon2019calibration} for the different models across dataset splits. In the RA case, DT-S and DT-BLS yield the highest predictive performance, with DT-BLS performing best. The RNN performs comparably to the standard decision tree but is outperformed by the meta-models, highlighting the advantages of structure-aware models in this context.

\begin{figure*}[htbp]
\floatconts
  {fig:trees}
  {\caption{Decision trees from the meta-model described in Section~\ref{sec:framework}, fitted to post-baseline events in RA: (a) treatment prediction and (b) treatment switching prediction. In (a), each node specifies the probability of the most common treatment. }}
  {%
    \subfigure[Treatment prediction.]{\label{fig:fu_tree}%
      \includegraphics[width=0.495\linewidth]{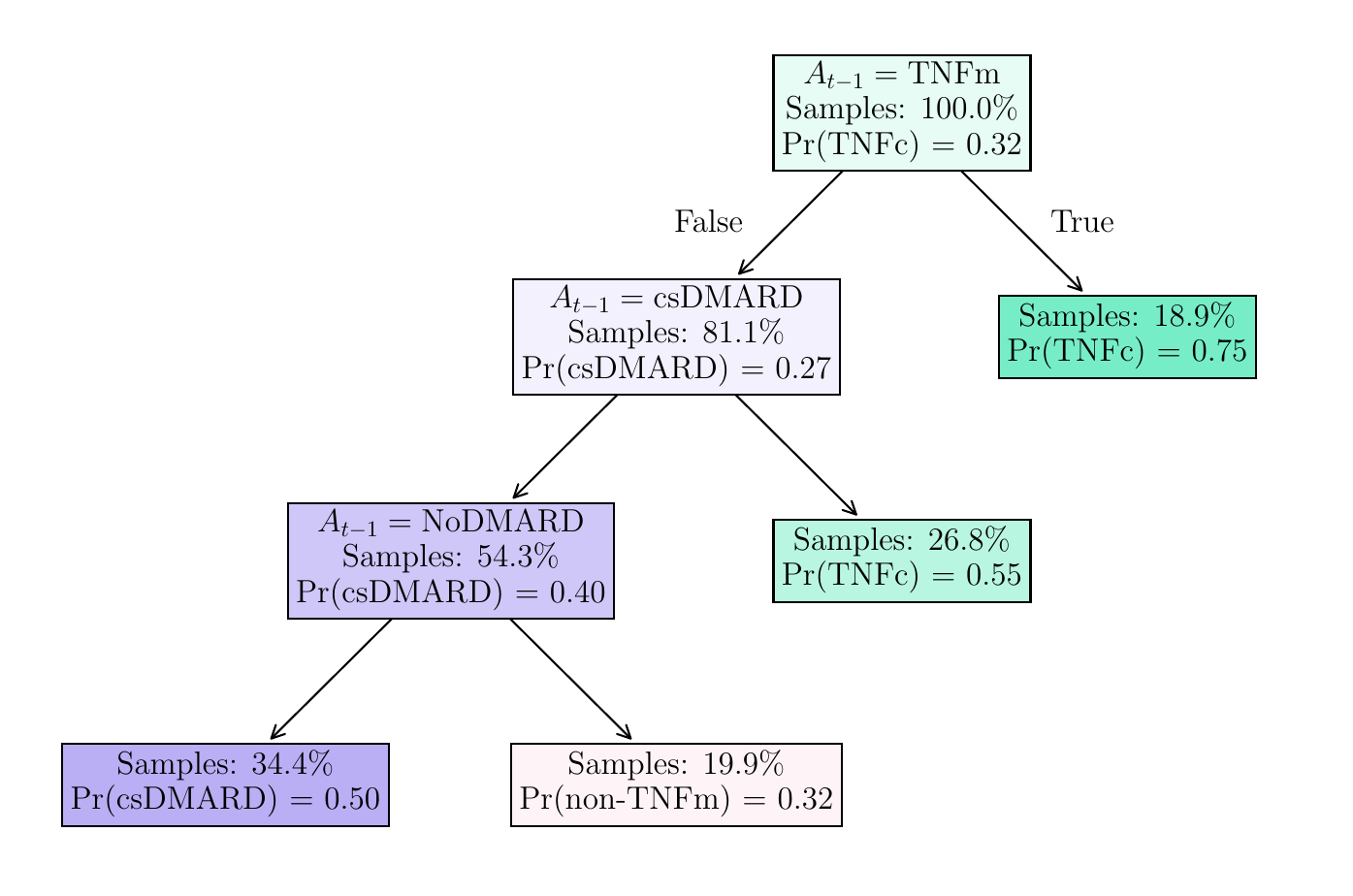}}%
    \hfill
    \subfigure[Treatment switching prediction.]{\label{fig:switch_tree}%
      \includegraphics[width=0.495\linewidth]{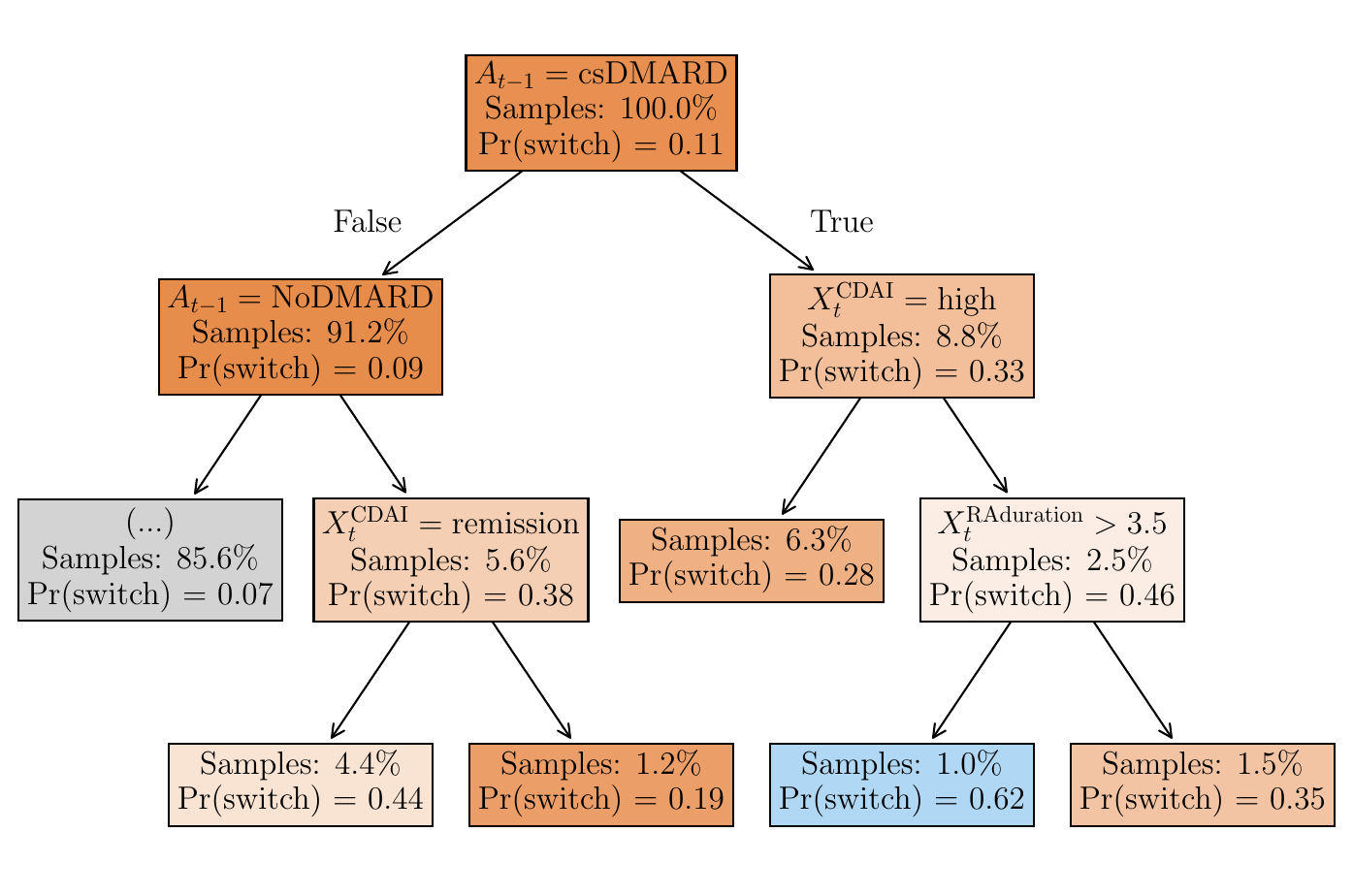}}
  }
\end{figure*}

For sepsis, all tree-based models exhibit similar performance, which is not surprising. Unlike RA, where treatment involves follow-up visits scheduled several months apart, sepsis management requires continuous administration of intravenous fluids and vasopressors. As a result, it cannot be as naturally decomposed into a two-step decision process—first deciding whether to switch treatment, then determining what treatment to switch to—as is possible with RA.

To examine the behavior policy for RA, we fit DT-BLS to the full cohort dataset (1,565 patients), with hyperparameter selection performed using 3-fold cross-validation. At baseline, TNF-based therapies are the most common, aligning with clinical guideline recommendations~\citep{smolen2023eular}. In Figure~\ref{fig:switch_tree}, we show the decision tree fitted to predict whether a patient switches treatment. The tree captures a common pattern described by \citet{matsson2024patterns}: patients often switch treatment after a period without DMARDs or while on a conventional synthetic (cs) DMARD therapy. In such cases, higher CDAI scores increase the probability of switching therapy. Among the large group of patients in the leftmost subtree, the probability of switching treatment is generally low ($<\SI{7}{\percent}$).

Figure~\ref{fig:fu_tree} shows the decision tree fitted to cases where RA patients switched therapy. Each node specifies the probability of the most common treatment, which the deterministic target policy from Section~\ref{sec:framework} would recommend in these situations. The previous treatment, $A_{t-1}$, provides a strong signal for determining which therapy patients switch to. Patients previously on TNF monotherapy or csDMARD therapy are likely to switch to TNF combination therapy. For patients previously without DMARD treatment, TNF monotherapy is the most common choice. These transitions are consistent with the patterns described in~\citet{matsson2024patterns}.

\subsection{Constructing and Evaluating Candidate Target Policies}
\label{sec:ope}

Next, we construct candidate target policies under the pragmatic policy development framework. We choose the DT-BLS behavior policy model, given its superior performance in the RA setting. We derive target policies based on the $k = \{1, 2, 3\}$ most common treatments (\verb|MC|), including variants that incorporate observed outcomes (\verb|MC+O|). When $k=1$, these policies are equivalent.

For comparison, we also include a policy that recommends a randomly selected treatment, as well as four policies learned using reinforcement learning: standard Q-learning (QL), deep Q-learning (DQN)~\citep{mnih2015human}, batch-constrained Q-learning (BCQ)~\citep{fujimoto2019off}, and conservative Q-learning (CQL)~\citep{kumar2020conservative}. Standard Q-learning is applied to a discrete Markov decision process based on clustered states, whereas DQN, BCQ, and CQL operate on continuous state representations. BCQ and CQL are specifically designed to address the distributional shift between behavior and target policies that arises in offline RL. See Appendix~\ref{app:exp_details} for further details.

In Table~\ref{tab:estimates_ess}, we present average OPE estimates of the policies using weighted importance sampling (WIS)~\citep{precup2000eligibility}, along with the corresponding average effective sample sizes (ESSs)~\citep{mcbook}. A small ESS relative to the number of trajectories indicates that a few weights dominate the weighted sum, limiting the reliability of the estimate. In RA, where the number of follow-up visits differs across patients, the estimated policy values are normalized accordingly.

\begin{table*}[t]
\small
\centering
\caption{The average value estimate $\hat{V}$ using weighted importance sampling (WIS) and effective sample size (ESS) for different target policies in RA and sepsis. The value of the behavior policy is estimated as the average reward in the data. The confidence intervals represent the interquartile range of each distribution. 
}
\label{tab:estimates_ess}
\begin{tabular}{lcccc}
\toprule
 & \multicolumn{2}{c}{\textbf{RA}} & \multicolumn{2}{c}{\textbf{Sepsis}} \\
 \cmidrule(lr){2-3} \cmidrule(lr){4-5}
Target policy & $\hat{V}_{\mathrm{WIS}}^{\pi}$ ($\uparrow$) & ESS ($\uparrow$) & $\hat{V}_{\mathrm{WIS}}^{\pi}$ ($\uparrow$) & ESS ($\uparrow$) \\
\midrule
\verb|MC| ($k=1$) & 0.7 {\scriptsize (0.4, 0.8)} & 406.1 {\scriptsize (388.1, 415.6)} & 74.1 {\scriptsize (66.8, 82.9)} & 64.1 {\scriptsize (46.1, 80.8)} \\
\verb|MC| ($k=2$) & 0.0 {\scriptsize ($-0.2$, 0.2)} & 566.1 {\scriptsize (553.3, 575.0)} & 74.6 {\scriptsize (72.0, 79.4)} & 277.7 {\scriptsize (250.6, 296.0)} \\
\verb|MC| ($k=3$) & $-0.5$ {\scriptsize($-0.6$, $-0.2$)} & 639.2 {\scriptsize(624.6, 650.3)} & 75.1 {\scriptsize(73.7, 76.6)} & 628.9 {\scriptsize(604.9, 647.7)} \\
\verb|MC+O| ($k=1$) & 0.7 {\scriptsize (0.4, 0.8)} & 406.1 {\scriptsize (388.1, 415.6)} & 74.1 {\scriptsize (66.8, 82.9)} & 64.1 {\scriptsize (46.1, 80.8)} \\
\verb|MC+O| ($k=2$) & 1.2 {\scriptsize (0.1, 2.7)} & 19.2 {\scriptsize (7.4, 31.7)} & 80.7 {\scriptsize (75.7, 93.9)} & 15.5 {\scriptsize (7.8, 24.7)} \\
\verb|MC+O| ($k=3$) & 3.2 {\scriptsize(0.8, 4.1)} & 17.1 {\scriptsize(9.0, 25.7)} & 85.3 {\scriptsize(68.5, 95.5)} & 6.9 {\scriptsize(3.0, 14.2)} \\
\midrule
RL (QL) & $0.0$ {\scriptsize ($-4.2$, 3.7)} & 3.0 {\scriptsize (2.1, 4.3)} & 86.0 {\scriptsize (80.3, 92.3)} & 14.0 {\scriptsize (6.8, 24.5)} \\
RL (DQN) & $0.0$ {\scriptsize ($-4.3$, 3.0)} & 5.3 {\scriptsize (2.4, 9.5)} & 89.3 {\scriptsize (70.9, 98.9)} & 1.7 {\scriptsize (1.1, 8.0)} \\
RL (BCQ) & $-1.8$ {\scriptsize ($-4.6$, 0.4)} & 7.3 {\scriptsize (4.4, 10.5)} & 83.0 {\scriptsize (76.3, 88.2)} & 19.3 {\scriptsize (12.7, 28.9)} \\
RL (CQL) & $-0.9$ {\scriptsize ($-5.3$, 1.2)} & 5.7 {\scriptsize (2.8, 11.4)} & 67.3 {\scriptsize (34.4, 85.2)} & 6.5 {\scriptsize (2.7, 11.5)} \\
\midrule
Random & 1.7 {\scriptsize ($-4.8$, 5.1)} & 1.3 {\scriptsize (1.0, 2.0)} & 97.9 {\scriptsize (59.7, 99.6)} & 1.6 {\scriptsize (1.1, 2.3)} \\
\midrule
Behavior policy & $-1.1$ {\scriptsize ($-1.2$, $-1.0$)} & 779.0 {\scriptsize (770.3, 788.8)} & 71.6 {\scriptsize (70.7, 72.2)} & 2297.0 {\scriptsize (2297.0, 2297.0)} \\
\bottomrule
\end{tabular}
\end{table*}

For both RA and sepsis, we find that the deterministic \verb|MC| policy ($k=1$) is, on average, estimated to outperform the behavior policy. In the RA case, the confidence intervals for the value estimates of these policies are well separated. In both settings, the average ESS for the \verb|MC| policies with $k=1$ is substantially larger than that of the RL-based policies (406.1 vs. 3.0–7.3 for RA, and 64.1 vs. 1.7–19.3 for sepsis). The small ESSs for the RL policy estimates, together with their high variance, underscore the challenges of reliable OPE and highlight the importance of keeping target policies close to the behavior policy.

For the \verb|MC| policies, adjusting $k$ provides a direct way to control overlap with the behavior policy and, consequently, the variance of the OPE estimates. This effect is particularly evident in the sepsis case, where increasing $k$ from 1 to 3 increases the ESS by a factor of 10. Interestingly, in the RA case, the variance remains largely unchanged when varying $k$ in the same way. For the \verb|MC+O| policies, which select the treatment with the highest estimated outcome among the $k$ most common options, reducing $k$ decreases the variance of the OPE estimates. While \verb|MC+O| policies show promise in terms of estimated value, the high variance in these estimates warrants cautious interpretation.

By normalizing the estimated target policy values relative to the behavior policy in RA, we can interpret them as the average change in CDAI per patient and treatment stage if these policies were to replace the behavior policy. Specifically, negative values indicate a decrease in CDAI. As shown in Figure~\ref{fig:ope_results}, the \verb|MC| policy suggests a potential decrease in CDAI for $k = 1, 2, \ldots, 8$, at which point the behavior policy is recovered.
As discussed above, the \verb|MC+O| policy indicates an even greater potential reduction in CDAI, though the high variance of the estimates limits their reliability.

\begin{figure}[t]
\floatconts
  {fig:ope_results}
  {\caption{OPE using weighted importance sampling for RA target policies based on the most common treatments under the behavior policy. Boxes represent the interquartile range of the value distribution, normalized relative to the behavior policy.}}
  {\includegraphics[width=0.65\linewidth]{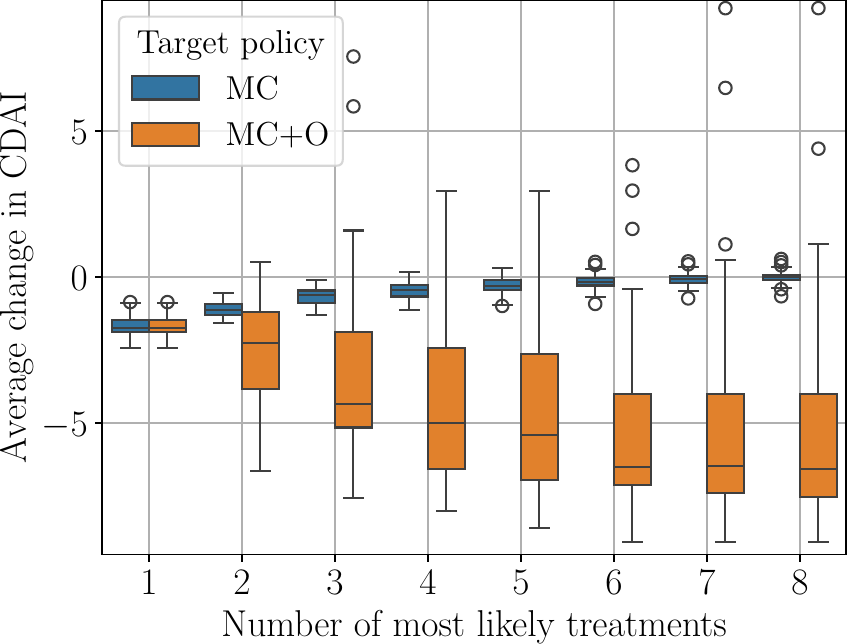}}
\end{figure}

\section{Discussion}

In this work, we proposed \emph{pragmatic policy development}, a framework for deriving target policies for clinical decision-making through interpretable behavior cloning. Specifically, to support both interpretability and off-policy evaluation, we constructed target policies based on the most commonly selected treatments in each patient state, as identified by a tree-based behavior policy model. To improve behavior policy accuracy, we designed the model to capture inherent data structure, such as patients’ tendency to remain on the same treatment across decision points. Our evaluation on real-world rheumatoid arthritis and sepsis treatment data demonstrated that interpretable policies based on the most common treatments offer a promising clinical approach, providing stronger statistical support in value estimates than reinforcement learning policies.

Intuitively, policies derived under this framework correspond to acting like the majority of clinicians in the observed data, leveraging collective expertise. Thus, such policies can be seen as standardizing frequent treatment patterns and reducing unwarranted practice variation—a common issue in healthcare~\citep{pace2022poetree}. While these policies may not represent the optimal strategies derivable from the data, their design ensures they can be evaluated meaningfully. After all, what cannot be evaluated has limited practical value. Additionally, the interpretability of these policies facilitates their use in clinical practice, where transparency is essential for validation and gaining trust from end users.

Our work has some limitations. First, we focused exclusively on off-policy evaluation using importance sampling. Doubly robust methods~\citep{jiang2016doubly, thomas2016data} could help reduce the variance of value estimates; however, all off-policy evaluation methods suffer to some extent from large discrepancies between target and behavior policies~\citep{voloshin2021empirical}. Second, we assumed that the state $S_t$ includes all confounding variables that causally affect both treatment and outcome. However, in practice, we were limited to the variables available in the observed data, meaning that unmeasured confounders could still be present. This assumption is particularly critical for the outcome-guided policy: if unmeasured confounders exist within the leaves of a fully grown tree, the estimated value of this policy may be biased upwards. Third, while previous research has shown that interpretable models such as sparse decision trees can effectively model policies for sequential decision-making~\citep{matsson2024how}, there may be cases where these models fit the data unsatisfactorily compared to more flexible models like neural networks. Prototype-based models may offer a better solution in such cases~\citep{ming2019interpretable,matsson2024how}.

To conclude, we summarize our contributions as three recommendations for researchers and practitioners:
\begin{enumerate}
    \item \textbf{Use interpretable models of the behavior policy.} Interpretable behavior policy models provide a basis for verifying assumptions and identifying candidate target policies that have sufficient support in the data.
    \item \textbf{Exploit known structure to improve behavior policy modeling.} If there are known patterns in the data—for example, differences in treatment assignment over time—models should be designed to account for these patterns, maintaining small model size while improving accuracy.
    \item \textbf{Construct target policies with statistical support under the behavior policy.} Evaluation is critical for turning observational data into actionable policies. Therefore, policies should be developed with evaluation feasibility in mind.
\end{enumerate}

\acks{This work was partially supported by the Wallenberg AI, Autonomous Systems and Software Program (WASP) funded by the Knut and Alice Wallenberg Foundation.

The computations and data handling were enabled by resources provided by the National Academic Infrastructure for Supercomputing in Sweden (NAISS), partially funded by the Swedish Research Council through grant agreement no. 2022-06725.}

\bibliography{references}

\appendix

\section{Datasets}
\label{app:data}

Our experimental evaluation was based on two distinct datasets related to the management of rheumatoid arthritis (RA) and sepsis. Here, we provide additional details on the datasets, including the full set of covariates used to represent patient states. Table~\ref{tab:datasets} summarizes key characteristics of these two decision-making processes.

\begin{table*}[t!]
\small
\centering
\caption{Characteristics of the decision-making processes studied in this work.}
\label{tab:datasets}
\begin{tabular}{lll}
\toprule
                                & RA                & Sepsis            \\
\midrule
Patients, n                     & 1,565             & 11,482            \\
Age in years, median (IQR)      & 59.0 (51.0, 67.0) & 66.0 (53.4, 77.8) \\
Female, n (\%)                  & 1,173 (75.1)       & 4996 (43.5)      \\
\midrule
Number of state variables $|\mathcal{S}|$   & 33                & 20                \\
Number of actions $|\mathcal{A}|$           & 8                 & 25                \\
Number of stages $T$, median (IQR)        & 4.0 (3.0, 5.0)    & 6.0 (6.0, 6.0) \\
\bottomrule
\end{tabular}
\end{table*}

\subsection{Rheumatoid arthritis}

RA is a chronic autoimmune disease that causes joint inflammation, pain, and potential deformity. It is primarily managed with disease-modifying antirheumatic drugs (DMARDs), which include conventional synthetic (cs), biologic (b), and targeted synthetic (ts) DMARDs. Biologic DMARDs are further divided into Tumor Necrosis Factor (TNF) inhibitors and non-TNF inhibitors. Clinical guidelines for RA management recommend starting treatment with a csDMARD upon clinical diagnosis~\citep{smolen2023eular}. If the initial treatment fails, adding a bDMARD or tsDMARD—typically a TNF inhibitor—is advised for patients with poor prognostic factors such as high disease activity.

Our original dataset comprised 42,068 unique patients enrolled in the PPD\texttrademark{} CorEvitas\texttrademark{} RA registry~\citep{corevitas}, a longitudinal clinical registry in the United States, between January 2012 and December 2021. According to the registry protocol, patients are recommended to complete follow-up visits every six months. We excluded records of visits where information related to therapy changes was missing, contradictory, or potentially erroneous, resulting in a cleaned dataset of 41,860 patients.\footnote{Only the subtrajectory up to the point of an excluded visit was considered for off-policy evaluation.} For example, visits where multiple b/tsDMARDs were prescribed simultaneously were excluded from the analysis, as such prescriptions are not clinically recommended and may represent a data reporting error.

We focused on RA treatment starting with the initiation of the first b/tsDMARD therapy, defined as the index visit (baseline). Specifically, we selected patients who (i) had no history of b/tsDMARD use before registry enrollment and (ii) initiated at least one b/tsDMARD therapy during registry participation. Sequences were truncated at visits where the subsequent visit lacked a clinical disease activity index (CDAI) measurement, necessary for reward computation. Additionally, follow-up visits were required to occur within intervals of 30 to 270 days, and patients needed a fixed follow-up duration of 2 years (720 days). Patients who did not meet these criteria were excluded from the cohort, resulting in 1,565 included patients.

Following \citep{matsson2024patterns}, we restricted the analysis to classes of DMARDs rather than individual drugs. Specifically, we divided bDMARDs into TNF inhibitors and non-TNF inhibitors, resulting in the following classes of drugs:  csDMARDs, TNF inhibitor biologics, non-TNF inhibitor biologics, and Janus kinase (JAK) inhibitors (the main group of tsDMARDs). We included both monotherapies and combinations of a csDMARD and a b/tsDMARD; for example, a TNF combined with a csDMARD. In total, the action space consisted of eight therapies ($K=8$): csDMARD therapy (\verb|csDMARD|), TNF biologic monotherapy (\verb|TNFm|), TNF biologic combination therapy (\verb|TNFc|), non-TNF biologic monotherapy (\verb|non-TNFm|), non-TNF biologic combination therapy (\verb|non-TNFc|), JAKi monotherapy (\verb|JAKm|), JAKi combination therapy (\verb|JAKc|), and no DMARD therapy (\verb|No| \verb|DMARD|).

We constructed the state using the variables listed in Table~\ref{tab:ra_variables}, which also provides descriptive statistics at baseline for each variable. Additionally, we included variables indicating historical comorbidities and previous treatment history, including the most recent therapy selection.

\begin{table*}[t!]
\small
\centering
\caption{Rheumatoid arthritis. A summary of variables included in the state representation and their baseline statistics. N represents the number of patients with non-missing baseline information.}
\label{tab:ra_variables}
\begin{tabular}{lll}
\toprule
Variable & N & Statistics \\
\midrule
Age in years, median (IQR) & 1,562 & 59 (51, 67) \\
RA duration in years, median (IQR) & 1,548 & 3 (1, 8) \\
Gender, n (\%) & 1,562 & \\
\qquad Male && 389 (24.9) \\
\qquad Female && 1,173 (75.1) \\
BMI, n (\%) & 1,535 & \\
\qquad Underweight && 14 (0.9) \\
\qquad Healthy weight && 356 (23.2) \\
\qquad Overweight && 492 (32.1) \\
\qquad Obesity && 673 (43.8) \\
Blood pressure, n (\%) & 1,561 & \\
\qquad Elevated && 247 (15.8) \\
\qquad Hypertension stage 1 && 531 (34.0) \\
\qquad Hypertension stage 2 && 390 (25.0) \\
\qquad Normal && 393 (25.2) \\
Currently pregnant, n (\%) & 1,041 & 3 (0.3) \\
Pregnant since last visit, n (\%) & 875 & 4 (0.5) \\
Private insurance & 1,565 & 1,131 (72.3) \\
Medicare insurance & 1,565 & 529 (33.8) \\
Medicaid insurance & 1,565 & 77 (4.9) \\
No insurance & 1,565 & 31 (2.0) \\
CDAI, n (\%) & 1,557 & \\
\qquad Remission && 175 (11.2) \\
\qquad Low && 370 (23.8) \\
\qquad Moderate && 514 (33.0) \\
\qquad High && 498 (32.0) \\
CCP outcome positive, n (\%) & 278 & 159 (57.2) \\
RF outcome positive, n (\%) & 304 & 191 (62.8) \\
Erosive disease, n (\%) & 1,183 & 98 (8.3) \\
Joint space narrowing, n (\%) & 368 & 218 (59.2) \\
Joint deformity, n (\%) & 361 & 65 (18.0) \\
Severe infections, n (\%) & 1,565 & 25 (1.6) \\
Tuberculosis outcome, n (\%) & 358 & 19 (5.3) \\
Comorbidities, n (\%) & 1,565 & \\
\qquad Metabolic diseases && 120 (7.7) \\
\qquad Cardiovascular diseases && 172 (11.0) \\
\qquad Respiratory diseases && 39 (2.5) \\
\qquad Cancer && 37 (2.4) \\
\qquad GI and liver diseases && 29 (1.9) \\
\qquad Musculoskeletal disorders && 475 (30.4) \\
\qquad Other diseases && 206 (13.2) \\
\bottomrule
\end{tabular}
\end{table*}

\subsection{Sepsis}

The dataset of patients with sepsis, defined according to the international Sepsis-3 criteria~\citep{singer2016third}, was sourced from the MIMIC-III database using code provided by \citet{komorowski2018artificial}. We focused on the first 24 hours of measurements, whereas the original work used 72 hours of data. Following~\citet{luo2024position}, we excluded patients with irregularly sampled data. For each patient, the data was structured as a multivariate time series with a discrete time step of 4 hours. We constructed the state using a subset of the variables from the original work (see Table~\ref{tab:sepsis_statics}). In addition to the variables in Table~\ref{tab:sepsis_statics}, we included the doses of intravenous fluids and vasopressors administered during the previous 4-hour period.

\begin{table*}[t!]
\small
\centering
\caption{Sepsis. A summary of variables included in the state representation and their baseline statistics. N represents the number of patients with non-missing baseline information.}
\label{tab:sepsis_statics}
\begin{tabular}{lll}
\toprule
Variable & N & Statistics \\
\midrule
Age in years, median (IQR) & 11,482 & 66.0 (53.5, 77.8) \\
Female, n (\%) & 11,482 & 4996 (43.5) \\
\midrule
Heart rate, median (IQR) & 11,482 & 87.8 (76.0, 101.0) \\
SysBP, median (IQR) & 11,482 & 118.0 (105.1, 133.9) \\
DiaBP, median (IQR) & 11,482 & 57.0 (48.3, 66.0) \\
MeanBP, median (IQR) & 11,482 & 77.2 (69.0, 87.2) \\
Shock index, median (IQR) & 11,482 & 0.7 (0.6, 0.9) \\
Hemoglobin, median (IQR) & 11,482 & 10.5 (9.3, 12.1) \\
Blood urea nitrogen, median (IQR) & 11,482 & 23.1 (15.0, 39.0) \\
Creatinine, median (IQR) & 11,482 & 1.1 (0.8, 1.7) \\
Total urine output, median (IQR) & 11,482 & 0.0 (0.0, 300.8) \\
Base excess, median (IQR) & 11,482 & 0.0 ($-2.6$, 3.0) \\
Lactate, median (IQR) & 11,482 & 1.7 (1.2, 2.6) \\
pH, median (IQR) & 11,482 & 7.4 (7.3, 7.4) \\
HCO3, median (IQR) & 11,482 & 24.0 (21.0, 27.0) \\
$\text{PaO}_{\text{2}}/\text{FiO}_{\text{2}}$ ratio, median (IQR) & 11,482 & 257.3 (167.5, 410.0) \\
Elixhauser, median (IQR) & 11,482 & 4.0 (2.0, 5.0) \\
SOFA, median (IQR) & 11,482 & 7.0 (5.0, 9.0) \\
\bottomrule
\end{tabular}
\end{table*}

\section{Experimental Details}
\label{app:exp_details}
 
We split each dataset into separate training and testing sets, with the former used to fit, validate, and calibrate the behavior policy model and the latter used for off-policy evaluation (OPE). For sepsis, we applied a standard 80/20 split. For RA, the data was divided equally between training and testing, and the cohort-selection criteria defined in the previous section were applied to both the testing and validation sets (\SI{20}{\percent} of the original training data in both cases). All training data was used to estimate the behavior policy in RA. However, the proportion of observations collected before the first b/tsDMARD initiation and the fraction of observations from patients who started b/tsDMARD therapy before registry enrollment were treated as hyperparameters in the modeling process. These adjustments were implemented due to the limited number of patients meeting the cohort criteria for RA.

The preprocessing steps differed between datasets. For sepsis, we followed the approach described in \cite{komorowski2018artificial}, standardizing normally distributed features and applying a log transformation followed by standardization for log-normally distributed features. For RA, missing values in numerical features were imputed using mean imputation, while categorical features were imputed using mode imputation and encoded using one-hot encoding. In both cases, the final model was calibrated using sigmoid calibration.

Below, we provide additional details for each experimental step: modeling the behavior policy, constructing target policy candidates, and evaluating each candidate using OPE. The entire procedure was repeated 50 times with different data splits to ensure statistical robustness of the results.

\subsection{Behavior Policy Modeling}{
We used cross-validation to select the best model for each of four model types, each chosen from 30 candidates generated by randomly sampling hyperparameters. The model types were: a standard decision tree (DT); a variant with separate trees for switch and treatment prediction (DT-S, described in Section~\ref{sec:structure}); an extension of DT-S with an additional tree for baseline treatment prediction (DT-BLS); and a recurrent neural network (RNN). Model selection was based on the area under the receiver operating characteristic curve (AUROC). We used scikit-learn's implementation for the decision trees, whereas the RNN was implemented in PyTorch.

Hyperparameters for DT and RNN are listed in Table~\ref{tab:hyperparams}. For DT-S and DT-BLS, we used the same hyperparameters as for DT for each decision tree in these meta-models. As noted above, in the RA setting, we introduced two additional hyperparameters: the proportion of observations collected before the first b/tsDMARD initiation, and the fraction of observations from patients who initiated b/tsDMARD therapy prior to registry enrollment. Each of these was evaluated at $\{0, 0.25, 0.5, 0.75, 1\}$. The RNN was trained for up to 100 epochs using cross-entropy loss and the Adam optimizer, with early stopping based on validation performance and a patience of 10 epochs.
}

\subsection{Target Policy Construction}{
In addition to the target policies based on the most commonly selected treatments under the behavior policy, we learned policies using different reinforcement learning methods: standard Q-learning (QL), deep Q-learning (DQN)~\citep{mnih2015human}, batch-constrained Q-learning (BCQ)~\citep{fujimoto2019off}, and conservative Q-learning (CQL)~\citep{kumar2020conservative}.

For the Q-learning approach, we followed \citep{komorowski2018artificial} by clustering the raw states using k-prototypes clustering for RA and k-means clustering for sepsis. For each case, we tested multiple values for the number of clusters and selected the solution with the lowest clustering loss. We then formulated a tabular Markov decision process (MDP) by assigning each raw state to its nearest cluster centroid and estimating transition and reward matrices from the observed data. Finally, we learned a target policy by solving the MDP with Q-learning, using a discount factor of 0.99. 

For the other methods---DQN, BCQ, and CQL---we used the implementation provided by~\citet{luo2024position}. To estimate the behavior policy, we employed a multi-layer perceptron for RA and a recurrent neural network for sepsis. This model was used to compute OPE estimates of the target policy during training, using truncated weighted importance sampling~\citep{luo2024position}. For both the behavior model and each target policy, we randomly sampled 10 sets of hyperparameters. We refer to \citet{luo2024position} for the full set of hyperparameters considered. In all cases, training was conducted over 50 epochs. All RL policies, as well as the random policy, were made deterministic in the RA setting. For the sepsis case, the policies were softened by assigning each action a \SI{1}{\percent} probability of being selected.
}

\subsection{Off-Policy Evaluation} 

We performed off-policy evaluation using weighted importance sampling (WIS)~\citep{precup2000eligibility}, defined as
\begin{equation*}
    \hat{V}^\pi_{\mathrm{WIS}} = \frac{\sum_{i=1}^n w_i \sum_{t=1}^{T}r_t^{(i)}}{\sum_{i=1}^n w_i} ,
    \label{eq:twis_value}
\end{equation*}
where the importance weights $w_i$ are defined as
\begin{equation*}
    w_i = \prod_{t=1}^{T} 
    \frac{p_\pi\left(A_t=a_t^{(i)} \mid S_t=s_t^{(i)}\right)}
    {p_{\hat{\mu}}\left(A_t=a_t^{(i)} \mid S_t=s_t^{(i)}\right)}.
    \label{eq:weight}
\end{equation*}
Unlike the standard importance sampling (IS) estimator, which normalizes by the number of evaluation trajectories $n$, the WIS estimator introduces bias but typically achieves lower variance.

To quantify how many samples meaningfully contribute to the value estimate, we computed the effective sample size $n_e~$\citep{mcbook}:
\begin{equation*}
    n_e = \frac{\left(\sum_{i} w_i\right)^2}{\sum_i w_i^2}.
    \label{eq:ess}
\end{equation*}

\begin{table*}[!t]
\small
\centering 
\caption{Hyperparameters of the decision tree (DT) and the recurrent neural network (RNN) along with their respective search spaces. For the meta-models DT-S and DT-BLS, each individual decision tree used the same hyperparameters as the standard DT. The same set of hyperparameters was applied to both RA and sepsis. However, for RA, we additionally considered two factors: (1) the proportion of observations collected before the first b/tsDMARD initiation and (2) the fraction of observations from patients who started b/tsDMARD therapy before registry enrollment. The following values were considered for each parameter: $\{0, 0.25, 0.5, 0.75, 1\}$.}
\label{tab:hyperparams}
\begin{tabular}{lll}
\toprule
Model & Hyperparameter & Values \\
\midrule
\multirow{1}{*}{DT}
& max depth & $\{2, 3, 4, 5, 6, 7, 8, 9\}$\\
& min fraction of samples per leaf & $\{0.01, 0.02, 0.03, 0.04, 0.05\}$\\
\midrule
\multirow{3}{*}{RNN} 
& learning rate & $\{0.001, 0.01, 0.1\}$ \\
& batch size & $\{32, 64, 128\}$ \\
& encoding dimension & $\{16, 32, 64\}$ \\
\bottomrule
\end{tabular}
\end{table*}

\section{Adjusting the Probability of Treatment Switching}
\label{app:switch_adjustment}

Using a separate model to predict whether a patient switches treatment—as described in the meta-model in Section~\ref{sec:structure}—allows us to adjust the probability of switching when constructing target policy candidates. Specifically, let $\bar{p}_{\hat{\mu}}^{s} \coloneqq p_{\hat{\mu}}^{s} + p_{1}$, where $p_1\in\left[-p_{\hat{\mu}}^{s}, 1-p_{\hat{\mu}}^{s}\right]$, be the adjusted probability of switching treatment. For example, by setting $p_1>0$, we can create target policies that more strongly encourage treatment switching compared to current practice. Note that we would still consider the most commonly selected treatments (see Equation~\eqref{eq:target}), but with an increased probability of selecting a different treatment.

Adjusting the probability of treatment switching in this way may lead to a target policy that partially violates the overlap assumption. Specifically, this means that $p_\pi\left(a \mid s\right) > 0$ when $p_\mu\left(a \mid s\right) \approx 0$ for some state-action pairs. In such cases, an IS-based OPE estimator could suffer from high variance, hindering accurate and reliable value estimation. Therefore, we recommend examining the modified switch probabilities for each leaf of the tree used to predict treatment switching: do they make clinical sense for each group of patients?

In Figure~\ref{fig:ope_results_switch}, we examine the effect of increasing the probability of switching therapy in RA. Specifically, we vary the parameter $p_1$ from 0 to 0.5. As shown in the figure, this results in a decrease in CDAI for most values of $k$, suggesting that increasing treatment switching---when applied across patient groups defined by the behavior policy model---may be a promising strategy for RA treatment. However, as shown in the right panel, the variance of the estimates increases substantially with higher values of $p_1$, indicating that these results should be interpreted with caution.

\begin{figure}[t]
\floatconts
  {fig:ope_results_switch}
  {\caption{Off-policy evaluation of target policies based on the $k$ most common treatments selected under the behavior policy, with the probability of treatment switching adjusted by adding a constant $p_1$. Value estimates are normalized relative to the behavior policy, representing the average decrease in CDAI per treatment stage if each target policy were used in place of the behavior policy.}}
  {\includegraphics[width=0.9\linewidth]{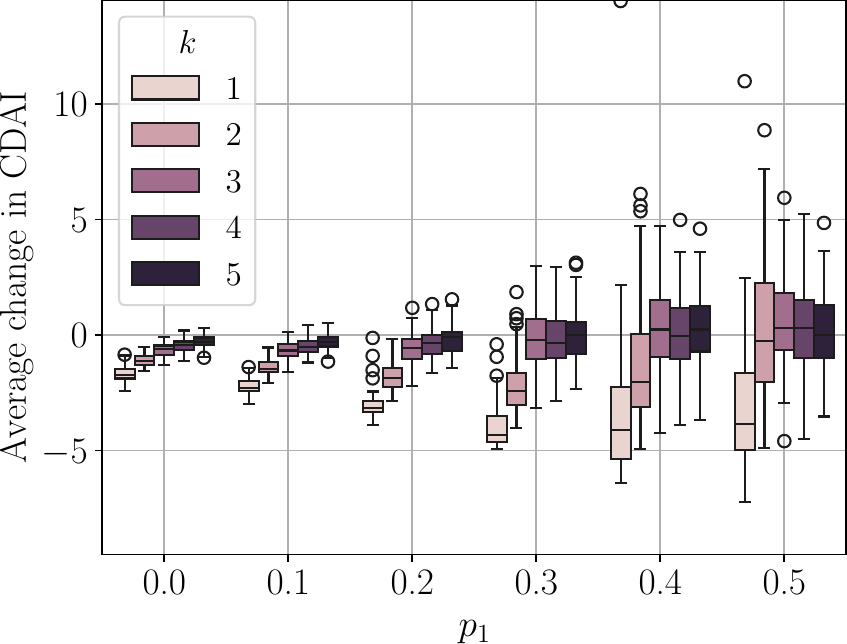}}
\end{figure}

\end{document}